\pdfoutput=1

\documentclass[numbers]{article}  

\usepackage[final]{ACL2023}

\usepackage{times}
\usepackage{latexsym}
\usepackage{balance}

\usepackage[T1]{fontenc}

\usepackage[utf8]{inputenc}

\usepackage{microtype}

\usepackage{inconsolata}

\usepackage{graphicx}
\usepackage{multirow}  
\usepackage{amssymb}

\usepackage{amsmath}
\usepackage{booktabs}  
\usepackage[noabbrev,capitalize]{cleveref}

\usepackage{colortbl}
\newcommand{\headercolor}{\rowcolor{gray!15}}

\usepackage{amsmath,amsfonts,bm}

\def\eqref#1{equation~\ref{#1}}

\def\1{\bm{1}}

\def\vd{{\bm{d}}}

\def\vq{{\bm{q}}}

\def\vs{{\bm{s}}}
\def\vt{{\bm{t}}}

\DeclareMathAlphabet{\mathsfit}{\encodingdefault}{\sfdefault}{m}{sl}
\SetMathAlphabet{\mathsfit}{bold}{\encodingdefault}{\sfdefault}{bx}{n}

\def\sD{{\mathbb{D}}}

\newcommand{\Ls}{\mathcal{L}}

\newcommand{\KL}{D_{\mathrm{KL}}}

\title{Unveil: Unified Visual-Textual Integration and Distillation for \\ Multi-modal Document Retrieval}

\author{
Hao Sun\textsuperscript{1,2$*$}, 
Yingyan Hou\textsuperscript{3,4}\thanks{$\quad$The first two authors contributed equally.}, 
Jiayan Guo\textsuperscript{1,2}\thanks{$\quad$Corresponding Authors.}\\
\textbf{
Bo Wang\textsuperscript{5},
Chunyu Yang\textsuperscript{6}, 
Jinsong Ni\textsuperscript{6}, 
Yan Zhang\textsuperscript{1,2}\textsuperscript{$^{\dag}$}}
\\
\textsuperscript{1}State Key Laboratory of General Artificial Intelligence, Peking University, Beijing, China\\
\textsuperscript{2}School of Intelligence Science and Technology, Peking University\\
\textsuperscript{3}Aerospace Information Research Institute, Chinese Academy of Sciences\\
\textsuperscript{4}Key Laboratory of Target Cognition and Application Technology\\
\textsuperscript{5}Beijing Institute of Technology,
\textsuperscript{6}Ucap Cloud
}

\begin{document}
\maketitle
\begin{abstract}
Document retrieval in real-world scenarios faces significant challenges due to diverse document formats and modalities.
Traditional text-based approaches rely on tailored parsing techniques that disregard layout information and are prone to errors, while recent parsing-free visual methods often struggle to capture fine-grained textual semantics in text-rich scenarios.
To address these limitations, we propose \textbf{Unveil}, a novel visual-textual embedding framework that effectively integrates textual and visual features for robust document representation. Through knowledge distillation, we transfer the semantic understanding capabilities from the visual-textual embedding model to a purely visual model, enabling efficient parsing-free retrieval while preserving semantic fidelity.
Experimental results demonstrate that our visual-textual embedding method surpasses existing approaches, while knowledge distillation successfully bridges the performance gap between visual-textual and visual-only methods, improving both retrieval accuracy and efficiency.
\end{abstract}

\section{Introduction}
\begin{figure}[t]
    \centering
    \includegraphics[width=.9\linewidth]{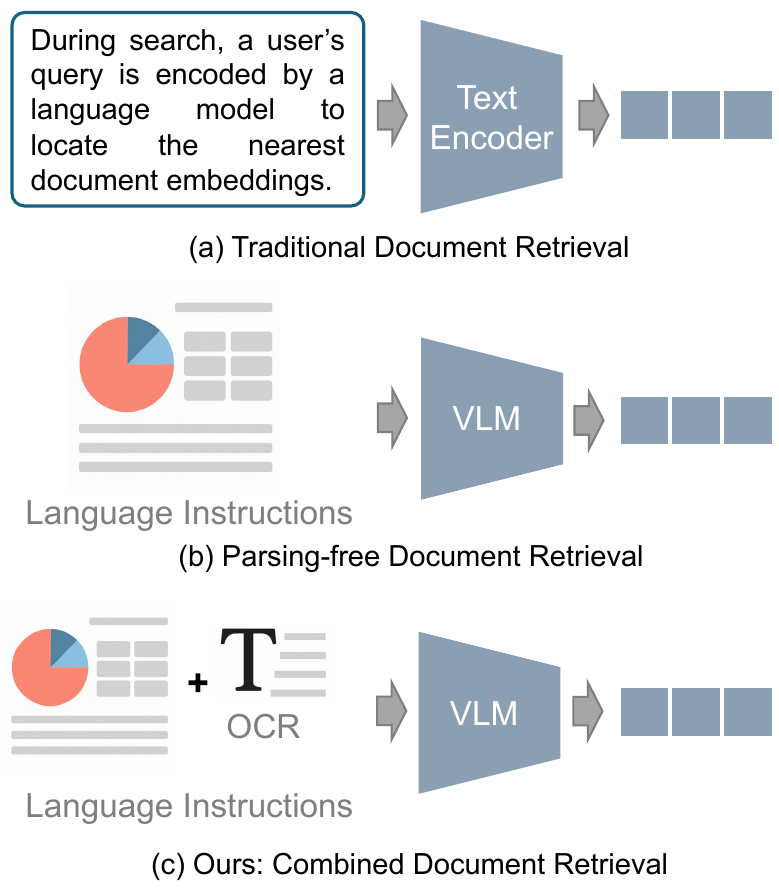}
    \caption{Comparison of document retrieval methods. Traditional approaches parse text and use text encoders for embeddings, while parsing-free methods directly process document screenshots with visual language models. Our visual-textual approach leverages both modalities, effectively addressing diverse scenarios.}
    \label{fig:comparison_of_methods}
\end{figure}

Document retrieval for real-world applications remains a challenging task due to the need to effectively handle diverse document formats, including text, images, charts, and complex visual layouts. As shown in \cref{fig:comparison_of_methods}, traditional document retrieval predominantly relies on Optical Character Recognition (OCR) to convert scanned or image-based documents into machine-readable text. Subsequently, approaches such as the lexical-based BM25 \cite{robertson2009probabilistic} and embedding-based techniques like Dense Passage Retrieval (DPR) \cite{karpukhin2020dense} are utilized to model the semantic relevance between queries and documents.
However, OCR-dependent pipelines come with significant limitations. They not only add computational overhead but also introduce potential recognition errors. Furthermore, these approaches often miss crucial visual contextual elements, which are essential for comprehending document content~\cite{zhang2024document, faysse2024colpali, ma2024unifying}.

\begin{figure}[t]
    \centering
    \includegraphics[width=.9\linewidth]{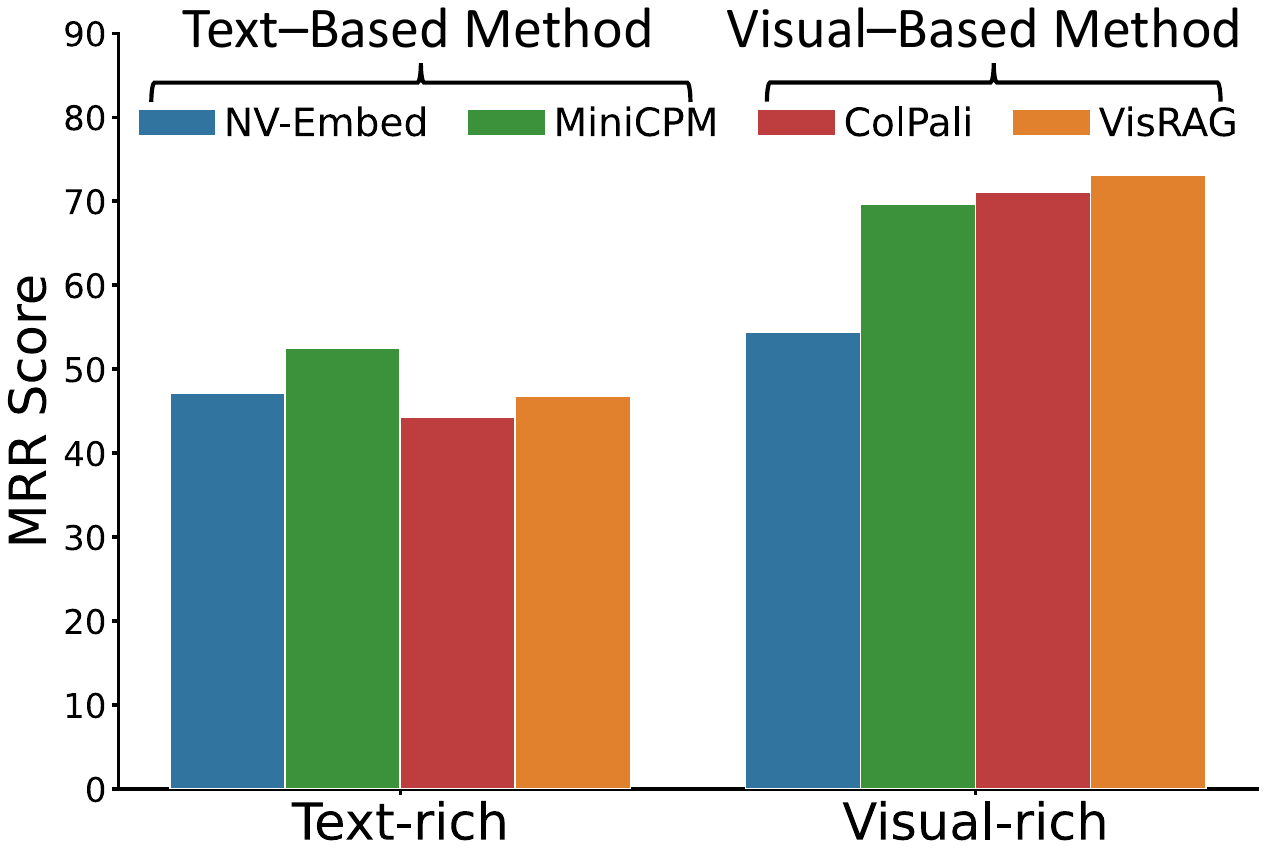}
    \caption{Empirical analysis on retrieval performance under text-rich and visual-rich scenarios.}
    \label{fig:empirical_study2}
\end{figure}

Recent research has shifted toward parsing-free techniques that directly utilize visual inputs such as document screenshots \cite{faysse2024colpali,ma2024unifying,zhou2024vista,ni2021large,lin2024mm,lee2024unified}. These methods leverage Vision-Language Models (VLMs) and Multi-Modal Large Language Models (MLLMs) to process entire pages directly, preserving rich structural and graphical information \cite{cho2024m3docrag, yu2024visrag, yao2024minicpmv}. While these approaches circumvent the computational overhead and complexity associated with OCR, our empirical analysis reveals significant limitations in their textual understanding capabilities.
As shown in \cref{fig:empirical_study2}, our comparison of text-based and visual-based methods across both text-rich scenarios (web-page retrieval) and visual-rich scenarios (visual document retrieval) reveals distinct performance patterns. In visual-rich scenarios where layout and graphical elements are crucial, these visual-based methods outperform traditional text-based approaches, highlighting their superior ability to process spatial and structural information \cite{chartqa, slidevqa, mpdocvqa}.
However, when handling text-rich contexts, visual-based methods struggle to capture semantic details that text-based methods process effectively.

This observation underscores a fundamental challenge: text-based methods excel in modeling linguistic semantics but overlook crucial layout and graphical details, while purely visual methods preserve visual context but struggle with fine-grained language understanding \cite{faysse2024colpali, zhang2024document, ni2021large, ma2024unifying}.
To address this limitation, we propose \textbf{Unveil} (\textbf{Un}ified \textbf{V}isual-T\textbf{e}xt \textbf{I}ntegration and Disti\textbf{l}lation), a novel framework that bridges the gap between textual and visual document understanding. Our approach consists of two key components:
First, we develop a visual-textual embedding approach that integrates both textual and visual inputs, leveraging the complementary strengths of both modalities for comprehensive document representations.
Second, we conduct knowledge distillation to transfer semantic understanding from the teacher model~(visual-textual embedding model) to the student model~(purely visual model), enabling enhanced text comprehension without OCR dependency during inference.
Specifically, we propose several techniques to facilitate this distillation process:
(1) Representation Alignment: The student model is trained to replicate the teacher model's representations by minimizing the distance between their query and document representations.
(2) Soft Label Distillation: We utilize the teacher model to provide a fine-grained label distribution for the student model.
(3) Adaptive Re-Weighting: We dynamically identify instances where discrepancies exist between the teacher and student models, assigning higher weights to these instances.

Our framework offers a flexible retrieval system. For text-rich scenarios that require precise semantic nuances, the visual-textual model—which incorporates both textual and visual inputs—can be employed. 
Alternatively, in scenarios where efficiency or OCR-free processing is preferred, the distilled visual-only model serves as a practical alternative while maintaining comparable semantic understanding.
We validate our approach on 12 datasets encompassing both text-rich and visual-rich scenarios. Experimental results indicate that the visual-textual embedding model consistently outperforms both text-based and visual-based methods. Furthermore, the distillation process effectively reduces the gap between visual-textual and visual-only approaches, enhancing retrieval accuracy and efficiency.

In summary, our contributions are as follows:
\begin{itemize}
    \item We identify that existing text-based and visual-based methods struggle to adapt across different scenarios. To address this, we introduce Unveil, a visual-textual embedding approach that integrates textual and visual features for comprehensive document understanding.
    \item We propose several knowledge distillation strategies to transfer the visual-textual model’s robust textual understanding to a purely visual model, enabling parsing-free retrieval without compromising accuracy.
    \item Extensive experiments demonstrate that our visual-textual embedding method outperforms existing text-based and visual-based methods. Additionally, the knowledge distillation effectively reduces the gap between visual-textual and visual-only approaches, enhancing retrieval accuracy and efficiency.
\end{itemize}

\begin{figure*}[htbp]
    \centering
    \includegraphics[width=.95\linewidth]{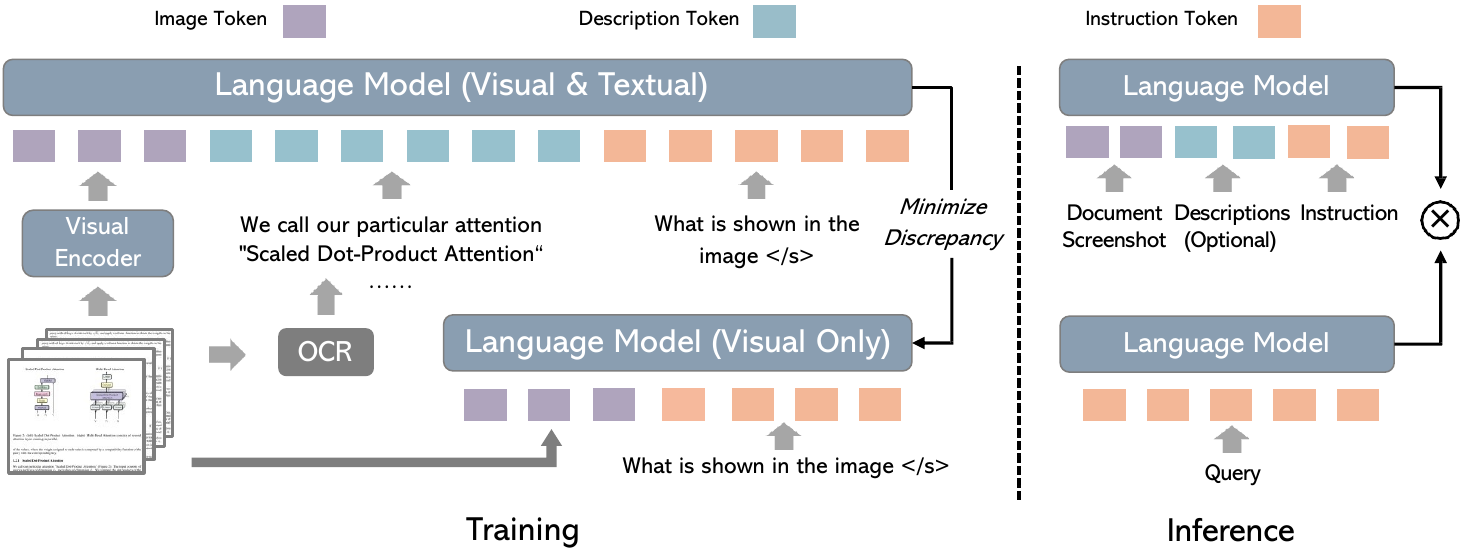}
    \caption{Unveil consists of: (a) a visual-textual embedding model that jointly processes document images and OCR text, and (b) a purely visual model that operates on document images only. During training, knowledge distillation is employed to transfer semantic understanding from the teacher (visual-textual) to the student (visual-only) model. At inference time, the framework offers flexibility to choose between the two models based on efficiency requirements.}
    \label{fig:framework}
\end{figure*}

\section{Methodology}
\label{sec:method}
Our proposed method seeks to bridge the gap between visual-textual and visual-only approaches. Initially, Unveil learns a textual-visual embedding model that leverages both OCR-derived text and visual inputs. Subsequently, it distills the strong capacity of the visual-textual embedding model into a purely visual model. The distilled visual model thus retains the semantic richness characteristic of visual-textual embeddings while achieving high efficiency without the need for textual input.

\subsection{Task Definition}
Given a query $q$ and a corpus $\mathcal{C}$ comprising visual documents $\{d_1, d_2, \ldots, d_n\}$, the task of multi-modal document retrieval is to identify the $k$ visual documents most relevant to the query $q$. Relevance is assessed using a similarity metric to measure the similarity between the query and document embeddings. Here, a visual document represents a complete information snippet (e.g., a web article or a PPT page), while the query is purely textual.

\subsection{Unveil Framework}
As illustrated in Figure~\ref{fig:framework}, Unveil comprises two components: the teacher model (Visual-Textual model) and the student model (Visual-Only model). We initially train these models independently, followed by knowledge distillation to produce a visual-only model capable of robust document retrieval without OCR reliance during inference.

\paragraph{Unified View of Retrieval Models}
Both the visual-textual and visual-only models employ a dual encoder architecture to model the similarity between queries and documents. The key distinction lies in the input to the document encoder.

For the visual-textual model's document encoder, we begin by employing OCR on each document image $d_i$ to derive a textual description $t_i$. The document image is then processed by the vision encoder of the vision-language model to yield visual tokens. The encoded visual latent embeddings are concatenated with a text prompt for input to the subsequent language model: \textit{``<s><img> <description> What is shown in this image?</s>''}.
In contrast, for the visual-only model's document encoder, the input to the subsequent language model is: \textit{``<s><img> What is shown in this image?</s>''}.
For both models, the input to the query encoder is the query text.
To aggregate sequence information using a language model with uni-directional attention, following prior work~\cite{ma2024unifying}, we use the embedding of the end-of-sequence token </s> from the last hidden state as the representation.
The representation of the queries and documents are calculated as follows:
\begin{equation}
\begin{split}
\vq &= \text{VLM}(q)[-1] \\
\vd_s &= \text{VLM}(\text{<img>}, \text{<inst>})[-1]\\
\vd_t &= \text{VLM}(\text{<img>}, \text{<desc>}, \text{<inst>}) [-1]
\end{split}
\label{equ:descore}
\end{equation}
where $\vd_s, \vd_t$ denote the document representation from the student and teacher, respectively.

The query-document similarity is measured using cosine similarity between their embeddings:
\begin{equation} 
\text{Sim}(\vq, \vd) = \frac{\vq^{T} \cdot \vd}{\|\vq\| \cdot \|\vd\|}
\label{equ:descore}
\end{equation}

During training, our embedding model is optimized using the InfoNCE loss \cite{oord2018representation}:
\begin{equation}
\begin{aligned}
\mathcal{L}_{\textrm{hard}} = -\sum_{d^+ \in \sD^+} \log\frac{\exp(\text{Sim}(\vq, \vd^+))}{\sum_{d \in \sD}\exp(\text{Sim}(\vq, \vd))}
\end{aligned}
\end{equation}

After independently training both models, we freeze the teacher model and leverage it to guide the student model during knowledge distillation.

\paragraph{Representation Alignment}
To align the representation of the student and teacher model, we define a representation alignment loss:
\begin{equation}
\mathcal{L}_{\text{align}} = \frac{1}{n} \sum_{i=1}^n (\| \vd_t - \vd_s \|_2^2 + \| \vq_t - \vq_s \|_2^2)
\end{equation}
where $n$ is the number of query-doc pairs.

Minimizing $\mathcal{L}_{\text{align}}$ encourages $\vd_s$ and $\vq_s$ to inherit the teacher’s textual representation abilities. As training progresses, the student model learns to encode in a manner reflecting both textual semantics and visual features, despite never explicitly encountering textual data during inference.

\paragraph{Soft Label Distillation}
The teacher model’s score distribution conveys fine-grained similarity information, unlike hard one-hot labels. We leverage this by aligning the student's distribution with the teacher's:
The label distribution of the student model and the teacher model are defined as follows:
\begin{align}
\vt &= \mathop{\forall}_{d \in\sD} \frac{\exp(\text{Sim}(\vq, \vd_t))}{\sum_{d' \in \sD}\exp(\text{Sim}(\vq, \vd'_t))}
\label{equ:studentscorepd} \\
\vs &= \mathop{\forall}_{d \in \sD} \frac{\exp(\text{Sim}(\vq, \vd_s))}{\sum_{d' \in \sD}\exp(\text{Sim}(\vq, \vd'_s))}
\end{align}

The soft label distillation loss is calculated as:
\begin{equation}
\Ls_{\textrm{soft}}=\KL(\vt/\tau, \vs/\tau)
\label{equ:ctdsoftloss}
\end{equation}
where $\tau$ is the temperature parameter.

\paragraph{Adaptive Re-Weighting}
Discrepancies between student and teacher models on certain documents can reveal student misinterpretations. We propose focusing on these discrepancies by giving them higher weights using KL Divergence:
\begin{equation}
w_i = \frac{\exp(\KL(t_i, s_i) / \tau)}{\sum_{j = 1}^K \exp(\KL(t_j, s_j) / \tau)}
\label{layer_selection}
\end{equation}
where $w_i$ denotes the importance of document $d_i$.

Finally, the total loss combines both the alignment loss and soft label distillation loss:
\begin{equation}
\Ls_{\textrm{total}} = \sum_{i=1}^{n} w_i \times (\Ls_{\textrm{align}}^i + \Ls_{\textrm{soft}}^i)
\end{equation}

\paragraph{Inference}
During inference, Unveil offers two inference modes to cater to different needs regarding performance and computational efficiency. 

The first mode, \textbf{Visual-Textual Mode}, uses both OCR text and document images to achieve optimal retrieval performance. By combining rich visual features with extracted textual information, it maximizes semantic understanding for scenarios requiring high precision.
The second mode, \textbf{Visual-Only Mode}, relies solely on the distilled visual model without OCR dependency, maintaining competitive accuracy through advanced visual representations. This mode significantly reduces computational overhead, making it ideal for efficiency-critical applications.


\section{Experiment Setup}
We evaluate Unveil on two distinct multi-modal retrieval scenarios: visual document retrieval, which emphasizes visual content, and web page retrieval, which focuses on textual content.

\subsection{Visual Document Retrieval}
\paragraph{Dataset}
We employ question-document pairs from various VQA datasets, each targeting distinct document types: MP-DocVQA~\cite{mpdocvqa} for industrial documents, ArXivQA~\cite{arxivqa}, ChartQA~\cite{chartqa}, InfographicsVQA~\cite{infographicvqa}, and PlotQA~\cite{plotqa} for different types of figures, as well as SlideVQA~\cite{slidevqa} for presentation slides. We adhere to the datasets' original train-test splits, except for MP-DocVQA and InfographicsVQA, where the validation split is utilized as our evaluation set. We construct the retrieval corpus by collecting the positive documents linked to each query from the training and evaluation sets.

\paragraph{Evaluation} 
Following conventional assessment approaches for VQA datasets, we apply Recall@10 and MRR@10 as evaluation metrics.

\begin{table*}[h]
\centering
\resizebox{1.0\textwidth}{!}{
\begin{tabular}{l|cccccccccccccc}
\toprule
\multicolumn{1}{c|}{\multirow{2}[3]{*}{\textbf{Model}}} & \multicolumn{2}{c}{\textbf{ArxivQA}} & \multicolumn{2}{c}{\textbf{ChartQA}} & \multicolumn{2}{c}{\textbf{DocVQA}} & \multicolumn{2}{c}{\textbf{InfoVQA}} & \multicolumn{2}{c}{\textbf{PlotQA}} & \multicolumn{2}{c}{\textbf{SlideVQA}} & \multicolumn{2}{c}{\textbf{AVG}} \\
\cmidrule(lr){2-3} \cmidrule(lr){4-5} \cmidrule(lr){6-7} \cmidrule(lr){8-9} \cmidrule(lr){10-11} \cmidrule(lr){12-13} \cmidrule(lr){14-15}
 & \textbf{Rec} & \textbf{MRR} & \textbf{Rec} & \textbf{MRR} & \textbf{Rec} & \textbf{MRR} & \textbf{Rec} & \textbf{MRR} & \textbf{Rec} & \textbf{MRR} & \textbf{Rec} & \textbf{MRR} & \textbf{Rec} & \textbf{MRR} \\
\midrule
\headercolor
\multicolumn{15}{c}{\textbf{Text-Based Models}} \\
BM25 & 42.30 & 32.48 & 56.69 & 43.66 & 86.38 & 73.56 & 83.19 & 69.03 & 50.48 & 33.18 & 91.16 & 76.65 & 68.37 & 54.76  \\
GTR & 40.82 & 31.17 & 56.55 & 43.50 & 74.19 & 57.09 & 84.95 & 67.82 & 44.87 & 28.83 & 90.43 & 74.82 & 65.30 & 50.54  \\
BGE-Large &38.40 & 29.78 & 53.76 & 42.47 & 77.54 & 59.63 & 87.98 & 70.86 & 47.60 & 32.06 & 92.26 & 75.77 & 66.26 & 51.76   \\
NV-Embed &  44.92 & 35.13 & 52.09 & 43.04 & 80.26 & 60.50 & 91.84 & 77.05 & 47.67 & 31.55 & 93.90 & 78.83 & 68.45 & 54.35   \\
MiniCPM & 69.53 & 57.02 & 73.96 & 60.91 & 93.24 & 80.56 & 94.67 & 82.34 & 63.86 & 45.07 & 96.93 & 92.30 & 82.03 & 69.70  \\
\midrule
\headercolor
\multicolumn{15}{c}{\textbf{Visual-Based Models}} \\
SigLIP & 50.50 & 35.57 & 66.16 & 47.62 & 54.55 & 34.86 & 68.08 & 47.40 & 52.82 & 25.89 & 87.22 & 78.06 & 63.22 & 44.90  \\
ColPali & 81.11 & 69.85 & 77.16 & 62.68 & 94.78 & 83.64 & 94.82 & 81.92 & 60.66 & 40.84 & 97.32 & 86.83 & 84.31 & 70.96 \\
DSE & 85.41 & 72.11 & 78.13 & 63.42 & 94.20 & 80.41 & 97.07 & 84.96 & 63.82 & 43.82 & 97.01 & 93.08 & 85.94 & 72.96   \\
VisRAG & 84.93 & 71.41 & 78.83 & 64.54 & 94.73 & 80.12 & 96.33 & 85.53 & 64.30 & 44.31 & 97.38 & 92.94 & 86.08 & 73.14  \\
\midrule
\headercolor
\multicolumn{15}{c}{\textbf{Hybrid Models}} \\
DSE & 77.57 & 63.76 & 74.51 & 62.68 & 93.88 & 81.55 & 96.58 & 85.39 & 64.22 & 45.70 & 97.20 & 93.84 & 83.99 & 72.16   \\
VisRAG & 83.58 & 69.48 & 77.58 & 64.35 & 95.64 & 83.27 & 96.63 & 85.70 & 64.61 & 46.00 & 97.71 & 93.61 & 85.96 & 73.74   \\
\midrule
\headercolor
\multicolumn{15}{c}{\textbf{Unveil (Ours)}} \\
Visual-Textual & \textbf{86.24} & \textbf{73.67} & 79.53 & \textbf{66.75} & \textbf{96.06} & \textbf{83.88} & 97.26 & 86.19 & 64.63 & 45.91 & \textbf{97.87} & \textbf{94.37} & 86.93 & \textbf{75.13} \\
Visual-Only & 86.23 & 73.27 & \textbf{80.36} & 66.40 & 95.74 & 82.53 & \textbf{97.61} & \textbf{86.39} & \textbf{64.82} & \textbf{46.16} & 97.61 & 93.75 & \textbf{87.06} & 74.75   \\
\bottomrule
\end{tabular}
}
\caption{Overall performance on Visual Document Retrieval. The best retrieval performance is marked in \textbf{bold}.}
\label{tab:visual_performance}
\end{table*}

\begin{table*}[h]
\centering
\resizebox{1.0\textwidth}{!}{
\begin{tabular}{l|cccccccccccccc}
\toprule
\multicolumn{1}{c|}{\multirow{2}[3]{*}{\textbf{Model}}} & \multicolumn{2}{c}{\textbf{NQ}} & \multicolumn{2}{c}{\textbf{TriviaQA}} & \multicolumn{2}{c}{\textbf{WebQ}} & \multicolumn{2}{c}{\textbf{2WikiMultihopQA}} & \multicolumn{2}{c}{\textbf{HotpotQA}} & \multicolumn{2}{c}{\textbf{ASQA}} & \multicolumn{2}{c}{\textbf{AVG}} \\
\cmidrule(lr){2-3} \cmidrule(lr){4-5} \cmidrule(lr){6-7} \cmidrule(lr){8-9} \cmidrule(lr){10-11} \cmidrule(lr){12-13} \cmidrule(lr){14-15}
 & \textbf{Rec} & \textbf{MRR} & \textbf{Rec} & \textbf{MRR} & \textbf{Rec} & \textbf{MRR} & \textbf{Rec} & \textbf{MRR} & \textbf{Rec} & \textbf{MRR} & \textbf{Rec} & \textbf{MRR} & \textbf{Rec} & \textbf{MRR} \\
\midrule
\headercolor
\multicolumn{15}{c}{\textbf{Text-Based Models}} \\
BM25 & 61.33 & 40.01 & 72.67 & 56.35 & 64.07 & 41.73 & 36.02 & 23.98 & 48.93 & 33.64 & 70.50 & 48.07 & 58.92 & 40.63    \\
GTR & 66.84 & 52.22 & 57.41 & 40.70 & 73.13 & 55.96 & 25.14 & 15.11 & 38.23 & 25.04 & 79.66 & 64.28 & 56.74 & 42.22   \\
BGE-Large & 68.39 & 54.22 & 61.03 & 44.36 & 73.97 & 56.99 & 26.65 & 16.74 & 42.55 & 29.05 & 80.67 & 66.46 & 58.88 & 44.64   \\
NV-Embed &  69.97 & 54.61 & 68.84 & 52.04 & 75.10 & 55.77 & 31.62 & 19.48 & 45.63 & 31.78 & 82.12 & 68.96 & 62.21 & 47.11   \\
MiniCPM  & 75.23 & 60.04 & 77.44 & 63.56 & 75.76 & 59.26 & 39.10 & 25.44 & 50.93 & 36.11 & 83.24 & 69.47 & 66.95 & 52.31 \\
\midrule
\headercolor
\multicolumn{15}{c}{\textbf{Visual-Based Models}} \\
SigLIP  &  59.57 & 41.45 & 53.25 & 34.08 & 58.33 & 39.09 & 22.30 & 13.93 & 31.63 & 18.89 & 67.82 & 47.11 & 48.82 & 32.42 \\
ColPali  &  68.78 & 53.18 & 60.70 & 44.82 & 73.57 & 56.50 & 27.59 & 16.97 & 40.77 & 27.43 & 81.68 & 66.47 & 58.85 & 44.23   \\
DSE  &  71.70 & 55.90 & 73.07 & 55.61 & 71.67 & 54.30 & 35.03 & 21.70 & 45.53 & 30.43 & 78.66 & 62.21 & 62.61 & 46.69  \\
VisRAG  &  72.17 & 56.20 & 72.37 & 55.55 & 71.38 & 53.96 & 34.13 & 20.46 & 45.60 & 31.06 & 79.78 & 63.72 & 62.57 & 46.83   \\
\midrule
\headercolor
\multicolumn{15}{c}{\textbf{Hybrid Models}} \\
DSE & 73.80 & 59.56 & 76.18 & 61.74 & 73.79 & 57.98 & 37.17 & 23.88 & 48.20 & 34.25 & 81.34 & 67.03 & 65.08 & 50.74  \\
VisRAG & 73.80 & 59.73 & 75.84 & 61.22 & 74.14 & 57.84 & 36.13 & 23.16 & 48.67 & 34.81 & 81.79 & 68.36 & 65.06 & 50.85  \\
\midrule
\headercolor
\multicolumn{15}{c}{\textbf{Unveil (Ours)}} \\
Visual-Textual  &  \textbf{75.80} & \textbf{61.86} & \textbf{78.75} & \textbf{64.68} & \textbf{75.81} & \textbf{60.02} & \textbf{40.57} & \textbf{26.23} & \textbf{52.40} & \textbf{36.62} & \textbf{84.13} & \textbf{70.40} & \textbf{67.91} & \textbf{53.30}  \\
Visual-Only &  72.20 & 57.30 & 74.64 & 59.07 & 73.84 & 55.84 & 35.50 & 22.23 & 48.13 & 32.95 & 80.11 & 65.31 & 64.07 & 48.78   \\
\bottomrule
\end{tabular}
}
\caption{Overall performance on Web-Page Retrieval. The best retrieval performance is marked in \textbf{bold}.}
\label{tab:web_performance}
\end{table*}

\subsection{Web-Page Retrieval}
\paragraph{Dataset}
Following~\cite{ma2024unifying}, we employ the Wiki-SS-corpus\footnote{\tiny \url{https://huggingface.co/datasets/Tevatron/wiki-ss-corpus}} as our retrieval corpus. This dataset is compiled from English Wikipedia pages via URLs, with screenshots captured automatically over four days, from May 20 to May 23, 2024. The corpus comprises 1,267,874 Wikipedia screenshots. To reduce inference time, we sample 112,888 screenshots to serve our retrieval corpus. For training, we use the Wiki-SS-NQ dataset\footnote{\tiny \url{https://huggingface.co/datasets/Tevatron/wiki-ss-nq}}, which is constructed by performing a BM25 search for each question to retrieve positive documents, thus forming query-document pairs.

Given the extensive use of the Wikipedia corpus in open-domain QA tasks, we make evaluation using several widely utilized QA datasets. These include open-domain QA datasets such as NQ \cite{kwiatkowski2019natural}, TriviaQA \cite{joshi2017triviaqa}, and WebQ \cite{berant2013semantic}, multi-hop datasets like 2WikiMultihopQA \cite{2wikimultihopqa-ho-2020} and HotpotQA \cite{yang2018hotpotqa}, as well as the ambiguous dataset ASQA \cite{asqa-stelmakh-2022}.

\paragraph{Evaluation} 
Consistent with previous practices for evaluating the effectiveness of retrieval in QA datasets, we use Recall@10 and MRR@10 as evaluation metrics.
Specifically, a question is considered correctly answered if its retrieved documents contain at least one answer from the answer list.

\subsection{Implementation Details}
Our framework involves initially training both a teacher and a student model independently, followed by knowledge distillation. Throughout both stages, models are fine-tuned using in-batch negatives for two epochs, with a batch size of 16 and a learning rate of 2e-5 on 8 NVIDIA A100 80GB GPUs. We initialize the models with MiniCPM-V 2.0~\citep{minicpmv2,yao2024minicpmv}. Additional details regarding the training and document parsing are provided in \cref{traing,parsing}.

\subsection{Baselines}
We compare our method with the following retrieval approaches:
\begin{itemize}
    \item \textbf{Text-Based Models}: This category encompasses BM25, a well-known lexical model, as well as advanced text embedding models such as BGE-Large-en-v1.5\footnote{\tiny \url{https://huggingface.co/BAAI/bge-large-en-v1.5}}\cite{bge_embedding}, GTR-T5-Large\footnote{\tiny \url{https://huggingface.co/sentence-transformers/gtr-t5-large}}\cite{ni2021large}, NV-Embed-v1\footnote{\tiny \url{https://huggingface.co/nvidia/NV-Embed-v1}}\cite{lee2024nvembed}, and MiniCPM\footnote{\tiny \url{openbmb/MiniCPM-V-2}}\cite{yao2024minicpmvgpt4vlevelmllm}, which has been fine-tuned for dense text retrieval.
    \item \textbf{Visual-Based Models}: This includes SigLIP\footnote{\tiny \url{https://huggingface.co/HuggingFaceM4/siglip-so400m-14-980-flash-attn2-navit}}\cite{zhai2023sigmoid}, a model in the CLIP style for vision tasks; ColPali\cite{faysse2024colpali}, a multi-vector retrieval model; as well as DSE \cite{ma2024unifying} and VisRAG \cite{yu2024visrag}, which are state-of-the-art visual embedding models.
    \item \textbf{Hybrid Models}: We also create hybrid models by interpolating similarity scores from the retrieval results of visual-based retriever like DSE and VisRAG and with text-based retrievers  MiniCPM \cite{ma2024fine}.
\end{itemize}

To ensure a fair comparison, all methods, except for widely used text-based methods, are trained on in-domain datasets.
Moreover, for the text-based methods, we trained MiniCPM on in-domain datasets to ensure a fair comparison.
\section{Experimental Results}
\subsection{Main Result}
In this section, we present experiments in both visual document retrieval and web page retrieval scenarios. Based on the results shown in \cref{tab:web_performance,tab:visual_performance}, several observations can be made:

First, text-based and visual-based models each exhibit unique advantages in different scenarios. For example, in web page retrieval, the text-based method MiniCPM significantly outperforms visual-based models. Conversely, in visual document retrieval, visual-based approaches excel. This highlights that these models cannot achieve superior performance across both scenarios. Interestingly, the simple lexical method BM25 outperforms more powerful dense retrieval models like BGE-Large. This can be attributed to the fact that text within these visual documents is often fragmented and semantically incoherent. In such cases, string matching might be a more effective solution.

Second, hybrid models yield intermediate results, which is understandable given that, in web-page retrieval, the scores generated by text-based models might be adversely affected by the less accurate scores from visual-based models, which leads to performance inferior to that of text-based models alone. This demonstrates that merely merging the outputs of the two models does not inherently enhance performance. Additionally, these models necessitate inference from both models, which increases the inference cost.

Third, our method Unveil, specifically the visual-textual variant, consistently achieves the highest performance across all retrieval scenarios, confirming its effectiveness in integrating information from both modalities for improved outcomes. Furthermore, the distilled visual-only version exhibits superior performance compared to both text-based and visual-based models and can even achieve performance comparable to the teacher model while requiring no text input.
This is mainly because our distillation framework can effectively transfer comprehensive knowledge to the student model.

\subsection{Ablation Study}
\begin{table}[t]
\centering
\resizebox{\linewidth}{!}{
\begin{tabular}{lcccc}
\toprule
 \multirow{2}{*}{\textbf{Methods}} & \multicolumn{2}{c}{\textbf{DocVQA}} & 
\multicolumn{2}{c}{\textbf{InfoVQA}}   \\
\cmidrule(lr){2-5} 
& \textbf{Rec} & \textbf{MRR} & \textbf{Rec} & \textbf{MRR}  \\
\midrule
Ours &  95.74 & 82.53 & 97.61 & 86.39  \\
-w/o Adaptive Re-Weighting  & 95.64 & 82.45 & 97.41 & 86.20  \\
-w/o Representation Alignment & 95.26 & 81.49 & 97.21 & 85.89  \\
-w/o Soft Label Distillation & 94.94 & 80.89 & 96.68 & 85.06  \\
-w/o Distillation & 94.20 & 80.41 & 97.07 & 84.96  \\
\bottomrule
\end{tabular}
}
\caption{Ablation Study. We experiment by gradually removing all components and observing the performance.}
\label{tab:ablation}
\end{table}
In this section, we evaluate the effectiveness of each component by incrementally removing them and observing the changes in performance. ``w/o  Representation Alignment'' and ``w/o  Soft Label Distillation '' refer to the removal of representation alignment and soft label distribution loss, respectively, following the removal of the adaptive re-weighting. ``w/o distillation'' represents the visual model before distillation.

As shown in \cref{tab:ablation}, removing each component results in performance degradation, confirming the effectiveness of each component. Specifically, we find that removing the representation alignment loss leads to significant degradation in model performance. This is because token representation contains the most valuable information about the query and document, and forcing the visual model to produce representations similar to the visual-textual model is the most direct way to learn from it. Additionally, removing the soft label distillation also results in performance degradation, primarily because the teacher provides a soft label that helps the student model discern fine-grained differences between documents within the same batch.

\subsection{Analysis}
\paragraph{Impact of Text Length}
\begin{table}[tbp]
\centering
\resizebox{1.0\linewidth}{!}{
\begin{tabular}{l|cccccc}
\toprule
 & \multicolumn{2}{c}{\textbf{NQ}} & \multicolumn{2}{c}{\textbf{TriviaQA}} & \multicolumn{2}{c}{\textbf{WebQ}} \\
 \cmidrule(lr){2-3} \cmidrule(lr){4-5} \cmidrule(lr){6-7}
\textbf{Length} & \textbf{Rec} & \textbf{MRR} &\textbf{Rec} & \textbf{MRR} &\textbf{Rec} & \textbf{MRR} \\
\midrule
\textbf{0}   & 71.70 & 55.90 & 73.07 & 55.61 & 71.67 & 54.30  \\
\textbf{512}   & 72.73 & 57.53 & 75.94 & 60.93 & 73.55 & 57.00   \\
\textbf{1024}   &  73.70 & 57.73 & 76.81 & 61.41 & 72.81 & 56.79  \\
\textbf{2048}   & 74.43 & 59.68 & 78.11 & 63.81 & 74.33 & 58.46  \\
\textbf{3096}   &  75.27 & 60.83 & 78.71 & 63.87 & 76.11 & 59.09  \\
\bottomrule
\end{tabular}
}
\caption{Performance of teacher model using different input text lengths.}
\label{tab:length}%
\end{table}%

In this section, we analyze the impact of text length on the performance of the teacher model. Specifically, we gradually increase the text length from 0 to 3096 and observe the changes in performance.

As shown in \cref{tab:length}, the model performance improves as the context length increases. This is expected because longer texts can provide more useful information about the image. However, longer contexts also incur additional inference costs, highlighting the importance of distilling the strong capabilities of the visual-textual teacher model into a visual-only student model.
Additionally, we observe a saturation phenomenon in performance. Specifically, there is a significant performance increase when the text length grows from 0 to 512, but the improvement becomes less pronounced as the length increases from 2048 to 3096. Therefore, selecting an intermediate text length might offer a good balance between effectiveness and efficiency.

\paragraph{Visualization of Embeddings}
\begin{figure}
\includegraphics[width=0.99\linewidth]{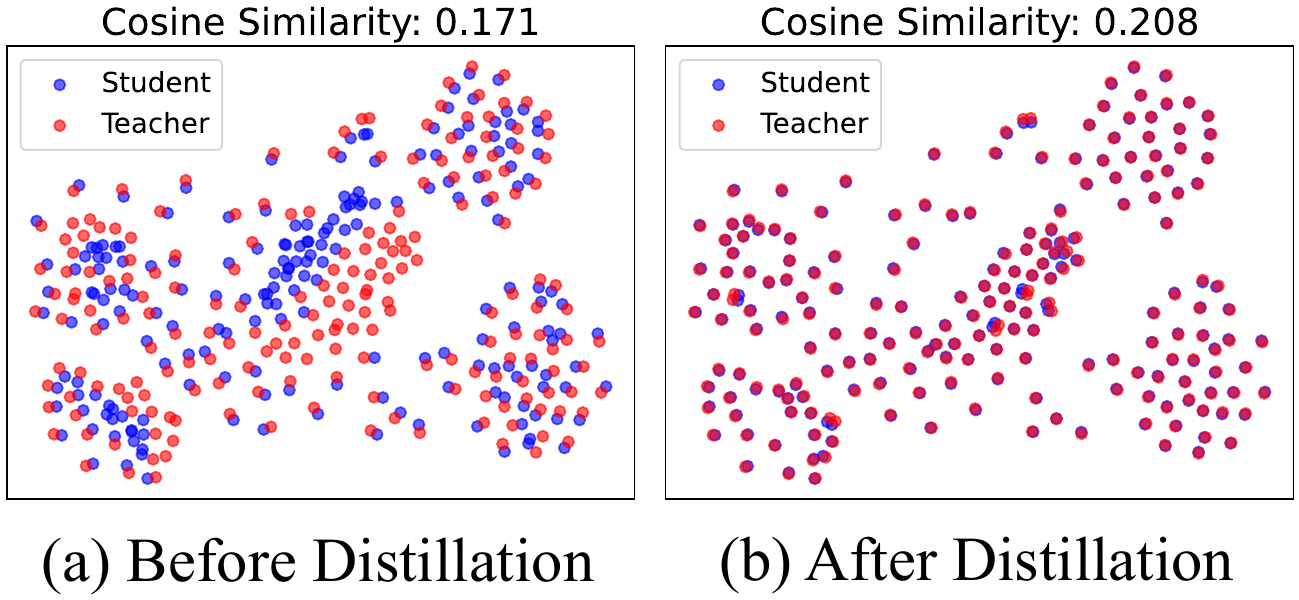}
\centering
\caption{The visualization of document embeddings.}
\label{fig:visualization}
\end{figure}

\begin{figure*}[htbp]
\includegraphics[width=\textwidth]{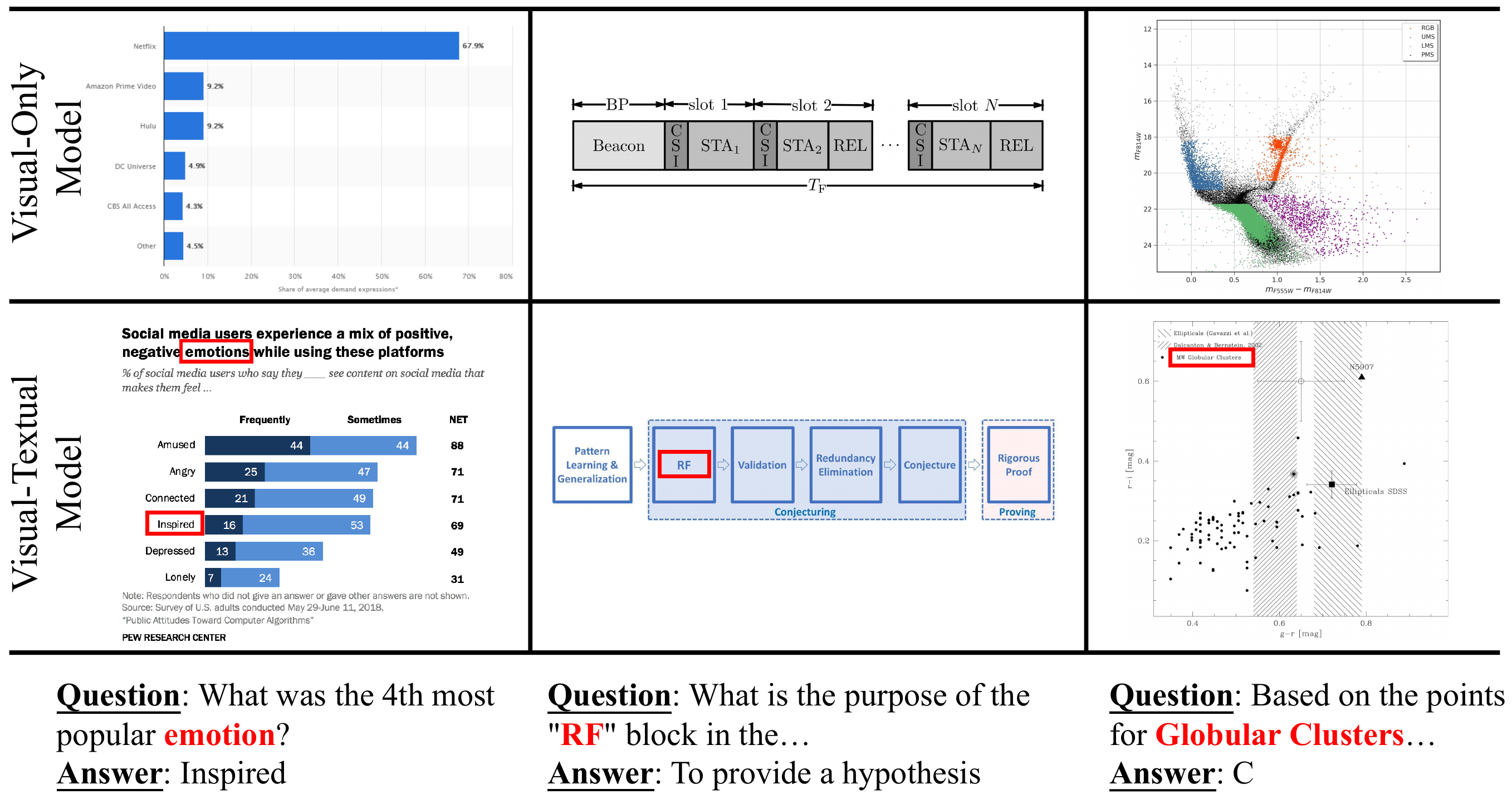}
\centering
\caption{Case Study. We sample several cases from the DocVQA and PlotQA datasets to compare the performance of the visual-only method and the visual-textual method. The keywords that match between the questions and the content of the images are highlighted in red.}
\label{fig:case}
\end{figure*}

In this section, we analyze the effects of the distillation process by visualizing the document representations of the student and teacher models before and after distillation. Specifically, we sample 200 documents from the ChartQA dataset and apply t-SNE to these document representations.

As shown in \cref{fig:visualization}, the representations of the student model become much more aligned with those of the teacher model after distillation, confirming the effectiveness of the representation alignment technique. Additionally, the cosine similarity between the student and teacher models also increases. Consequently, after distillation, the student model is able to achieve performance similar to that of the teacher model without incurring additional computational costs from input text.

\subsection{Case Study}
In this section, we analyze the effectiveness of the visual-textual model versus the visual-only model through several cases from the DocVQA and PlotQA datasets.

As illustrated in \cref{fig:case}, the visual-textual model demonstrates superior retrieval accuracy compared to its visual-only counterpart. The key distinction lies in their ability to capture semantic information: while the visual-only model primarily relies on visual patterns, potentially missing crucial textual context, the visual-textual model leverages both modalities to form a comprehensive understanding of the document content. This enhanced semantic comprehension directly translates to more accurate retrieval results.
\section{Related Work}

\subsection{Multi-modal Document Retrieval}

In multi-modal document understanding, models integrate visual and textual data to enhance information extraction. MPLUG-DocOwl \cite{ye2023mplug} introduces a modular multimodal large language model for OCR-free document understanding, leveraging both visual and textual content. MPLUG-DocOwl2 \cite{hu2024mplug} extends this approach by focusing on high-resolution compression for multi-page documents. VISTA \cite{zhou2024vista} offers a method for visualized text embedding, enabling efficient multi-modal retrieval across document types. Unified multi-modal representations, such as in \cite{lee2024unified}, combine text and image features for improved retrieval and understanding. Document parsing challenges are also addressed by recent work on structured information extraction \cite{zhang2024document}, which focuses on methods for extracting and understanding document structures.

\subsection{Multi-modal RAG}

Multi-modal retrieval-augmented generation (RAG) models combine retrieval and generative techniques, leveraging both textual and visual information to enhance document processing tasks. VisRAG \cite{yu2024visrag} uses vision-based retrieval to improve generative tasks such as document summarization by combining visual and textual content. M3DocRAG \cite{cho2024m3docrag} extends RAG to multi-page, multi-document settings, improving the generation of summaries and answers by incorporating information from multiple document sources. M-Longdoc \cite{chia2024m} introduces a retrieval-aware tuning framework that enhances the understanding of super-long documents by selecting relevant document segments for generation. Colpali \cite{faysse2024colpali} applies vision-language models with retrieval for more efficient document retrieval, thereby boosting the quality of generation tasks. MM-Embed \cite{lin2024mm} proposes a unified framework for multimodal retrieval with LLMs, optimizing retrieval and generation for multi-modal documents.
\section{Conclusion}
In this paper, we identify that current text-based and visual-based methods lack adaptability across different scenarios. To overcome this, we introduce Unveil, a visual-textual embedding approach that integrates text and visual document features for enhanced semantic grounding. Our knowledge distillation technique transfers robust textual understanding from the visual-textual model to a purely visual model, allowing for parsing-free retrieval without sacrificing accuracy. Empirical results show that our visual-textual embedding method surpasses existing text-based and visual-based approaches. Additionally, the knowledge distillation bridges the gap between visual-textual and visual-only models, improving retrieval accuracy and efficiency.
\section*{Limitations}
In this paper, we propose a multi-modal document retrieval framework that leverages both visual and textual information. We acknowledge a limitation in our approach: the visual-textual embedding model relies on textual inputs, necessitating OCR parsing of documents. This requirement can introduce additional computational overhead and may affect processing time, especially when dealing with large volumes of documents or when OCR accuracy is variable.

\section*{Acknowledgement}
This work is supported in part by Ucap Cloud and the State Key Laboratory of General Artificial Intelligence.
\section*{Ethics Statement}
This research was conducted in full compliance with the ACL Ethics Policy. All datasets and large language models (LLMs) used for evaluation purposes are publicly available, ensuring transparency and reproducibility of our results. Our work is aimed at advancing multi-modal embedding techniques to improve document retrieval capabilities. We have carefully considered ethical implications and do not foresee any negative ethical impacts arising from our research.


\bibliography{anthology,custom}
\bibliographystyle{acl_natbib}

\clearpage

\appendix
\onecolumn
\section{Dataset Statistics}
\begin{table*}[h]
\centering
\resizebox{\linewidth}{!}{
\begin{tabular}{lcccccc}
\toprule
\textbf{Settings} & \textbf{NQ} & \textbf{TriviaQA} & \textbf{WebQ} & \textbf{HotpotQA} & \textbf{2WikiMultihopQA} & \textbf{ASQA} \\
& \cite{kwiatkowski2019natural}  & \cite{joshi2017triviaqa} & \cite{berant2013semantic} & \cite{yang2018hotpotqa} & \cite{2wikimultihopqa-ho-2020} & \cite{asqa-stelmakh-2022} \\
\midrule
Task & Open-domain QA & Open-domain QA & Open-domain QA & Multi-hop QA & Multi-hop QA & Ambiguous QA \\
Test Data & 3,610 & 11,313 & 2,032 & 7,405 & 12,576 &  895\\
Metrics & \multicolumn{6}{c}{Recall@10, MRR@10} \\
\bottomrule
\end{tabular}
}
\caption{Statistics and experimental settings of different tasks/datasets.}
\label{tab:datasets}
\end{table*}

\begin{table*}[h]
\centering
\resizebox{\linewidth}{!}{
\begin{tabular}{lcccccc}
\toprule
\textbf{Settings} & \textbf{ArXivQA} & \textbf{ChartQA} & \textbf{MP-DocVQA} & \textbf{InfoVQA} & \textbf{PlotQA} & \textbf{SlideVQA} \\
& \cite{arxivqa}  & \cite{chartqa} & \cite{mpdocvqa} & \cite{infographicvqa} & \cite{plotqa} & \cite{slidevqa} \\
\midrule
Task & Arxiv Figures & Charts & Industrial Documents & Infographics & Scientific Plots & Slide Decks \\
Test Data &  8,640 & 718 & 1,879 & 2,046 &  11,307 & 1,640\\
Metrics & \multicolumn{6}{c}{Recall@10, MRR@10} \\
\bottomrule
\end{tabular}
}
\caption{Statistics and experimental settings of different tasks/datasets.}
\label{tab:datasets}
\end{table*}

\section{Training Details}
\label{traing}
\paragraph{Training Data}
In web page retrieval, we utilize 49,095 training pairs of query and positive documents.
In visual document retrieval, we utilize 122,752 training pairs of query and positive documents.

\paragraph{Training Process}
We conducted full parameter fine-tuning during both stages.
In the first stage, both student model and teacher were fine-tuned for 2 epochs with a learning rate of 2e-5 and a batch size of 16.
In the second stage, the teacher model was frozen and the student was fine-tuned for 2 epochs with a learning rate of 2e-5 and a batch size of 16.

\paragraph{Model Inference}  After fine-tuning on the web page retrieval dataset, we tested the model on all the open-domain datasets, including open-domain QA datasets such as NQ \cite{kwiatkowski2019natural}, TriviaQA \cite{joshi2017triviaqa}, and WebQ \cite{berant2013semantic}, multi-hop datasets like 2WikiMultihopQA \cite{2wikimultihopqa-ho-2020} and HotpotQA \cite{yang2018hotpotqa}, as well as the ambiguous dataset ASQA \cite{asqa-stelmakh-2022}.

After fine-tuning on the visual document retrieval dataset, we tested the model on all the visual document datasets, including  MP-DocVQA~\cite{mpdocvqa} for industrial documents, ArXivQA~\cite{arxivqa}, ChartQA~\cite{chartqa}, InfographicsVQA~\cite{infographicvqa}, and PlotQA~\cite{plotqa} for different types of figures, as well as SlideVQA~\cite{slidevqa}.

\section{Document Parsing}
\label{parsing}
Following \cite{yu2024visrag},
we use PaddlePaddle OCR (PPOCR)~\citep{du2020ppocr} for document parsing. The process involves several stages:

\begin{enumerate}
    \item \textbf{Text Detection}: A text detection model identifies text regions within the document and generates bounding boxes around them.

    \item \textbf{Orientation Classification}: These detected regions are processed by a classification model to correct any orientation issues, such as rotation or flipping.

    \item \textbf{Text Recognition}: A recognition model extracts the textual content from the corrected bounding boxes, returning the recognized text along with confidence scores. Only results with confidence scores above 0.6 are retained, and the bounding box coordinates, along with the recognized text, are stored for further processing.
\end{enumerate}

Throughout this process, we apply a Layout Preserving policy. This approach maintains the original document structure by ordering the text boxes based on their spatial positions. Spaces and line breaks are dynamically inserted to reflect horizontal and vertical gaps between text regions. This ensures that the extracted text mirrors the original document layout, preserving its formatting in the final output.

\end{document}